\documentclass[lettersize,journal]{IEEEtran}
\usepackage{amsmath,amsfonts}
\usepackage{algorithmic}
\usepackage{algorithm}
\usepackage{array}
\usepackage[caption=false,font=normalsize,labelfont=sf,textfont=sf]{subfig}
\usepackage{textcomp}
\usepackage{stfloats}
\usepackage{url}
\usepackage{verbatim}
\usepackage{graphicx}
\usepackage{cite}
\usepackage{amssymb}
\usepackage{lipsum}
\usepackage{amsfonts, amssymb, amsbsy, amsmath}
\usepackage{float}
\usepackage[export]{adjustbox}
\usepackage{siunitx} 
\usepackage{longtable}
\usepackage{booktabs}\usepackage{ragged2e}
\usepackage[table]{xcolor}

\begin{document}

\title{TADAP: Trajectory-Aided Drivable area Auto-labeling with Pre-trained self-supervised features in winter driving conditions}

\author{{Eerik Alamikkotervo, Risto Ojala, Alvari Seppänen, Kari Tammi}}
        



\maketitle

\begin{abstract}
Detection of the drivable area in all conditions is crucial for autonomous driving and advanced driver assistance systems. However, the amount of labeled data in adverse driving conditions is limited, especially in winter, and supervised methods generalize poorly to conditions outside the training distribution. For easy adaption to all conditions, the need for human annotation should be removed from the learning process. In this paper, Trajectory-Aided Drivable area Auto-labeling with Pre-trained self-supervised features (TADAP) is presented for automated annotation of the drivable area in winter driving conditions. A sample of the drivable area is extracted based on the trajectory estimate from the global navigation satellite system. Similarity with the sample area is determined based on pre-trained self-supervised visual features. Image areas similar to the sample area are considered to be drivable. These TADAP labels were evaluated with a novel winter-driving dataset, collected in varying driving scenes. A prediction model trained with the TADAP labels achieved a +9.6 improvement in intersection over union compared to the previous state-of-the-art of self-supervised drivable area detection.
\end{abstract}


\section{Introduction}
\IEEEPARstart{D}{detection} of the drivable area is a critical task for enabling safe autonomous driving. Currently, supervised deep learning methods dominate the field of visual detection, because they can learn advanced detection mechanisms from the training data. However, deep learning models are sensitive to changes caused by noise, occlusion, and weather effects \cite{feng2022survey} meaning they generalize poorly to conditions that are not included in the train set. Thus models trained in favorable weather, perform poorly in adverse conditions that include snow, rain, or fog. Winter conditions are particularly challenging if the model has not been trained in that domain. In winter, the boundary between drivable and non-drivable areas is not clearly defined, and changing snow conditions cause high variation in the drivable area appearance. For these reasons, models trained on favorable weather conditions can't produce reliable predictions in winter conditions. The same factors combined with a slippery road surface make winter conditions also demanding to human drivers. Snowy conditions are reported to increase the crash rate by up to 870\% \cite{strong2010safety}. Many countries experience at least short periods of snowfall and to implement autonomous driving and advanced driver assistance systems in these regions, detection of the drivable area must work in winter conditions.

As the drivable area appearance has more variation in winter, adaptability to changing conditions is important. Adaptability can be achieved with a self-supervised learning approach that doesn't require human supervision. There have been efforts to automate the learning process using self-supervision from depth sensors \cite{wang2019self,mayr2018self,cho2018multi,liu2018co,chen2023learning,shinzato2014road,shinzato2014roadlidar} but these approaches have difficulties in separating the drivable area from other flat areas like a sidewalk. Another popular approach is to learn the drivable area representations from the traveled trajectory \cite{schmid2022self,seo2023learning,lieb2005adaptive,tang2017one,seo2023scate,wellhausen2019should,zurn2020self} but the trajectory only captures a small part of the drivable area and learning to detect representations not included in the trajectory is challenging. However, in recent years, significant advances have been made in self-supervised feature learning \cite{caron2021emerging,zhou2021ibot,oquab2023dinov2,ilharco_gabriel_2021_5143773,he2022masked,caron2020unsupervised}, providing a potential solution. These models can learn meaningful visual features with no manual labels and can be trained with large diverse datasets, providing excellent generalizability.

\begin{figure}[t]
\setlength\extrarowheight{-4pt}
\begin{tabular}{@{}c@{}c@{}c@{}}
\centering

 & \footnotesize Previous & \footnotesize Model trained\\[0pt]
\footnotesize Input & \footnotesize state-of-the-art & \footnotesize with TADAP labels \\[0pt]
\includegraphics[width = 1.15in]{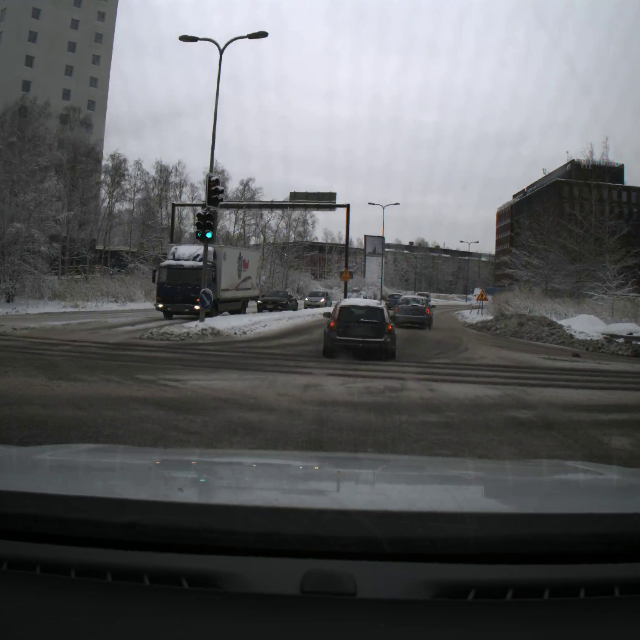} &
\vspace{0in}
\includegraphics[width = 1.15in]{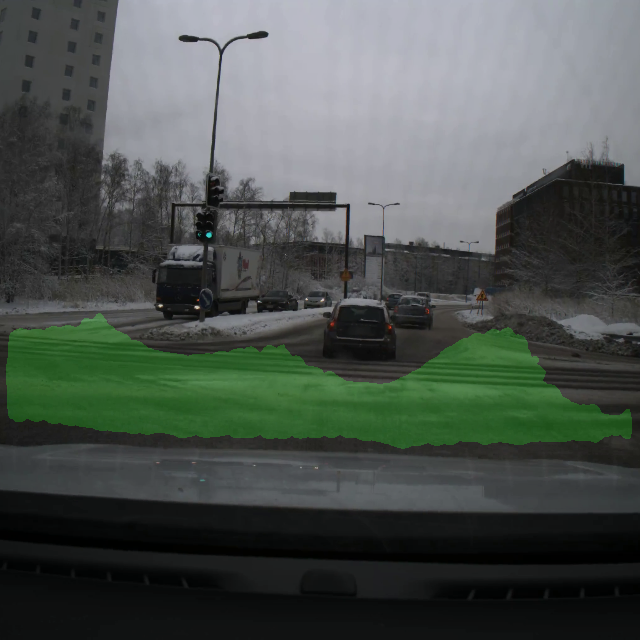} &
\vspace{0in}
\includegraphics[width = 1.15in]{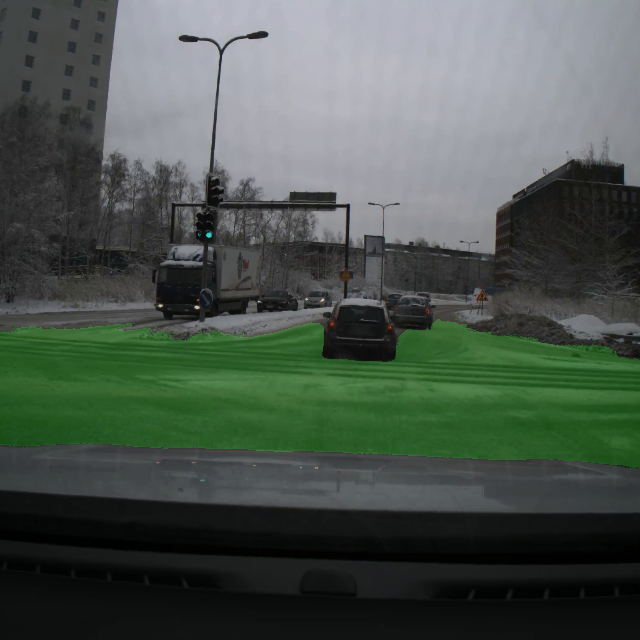}\\[0pt]
\includegraphics[width = 1.15in]{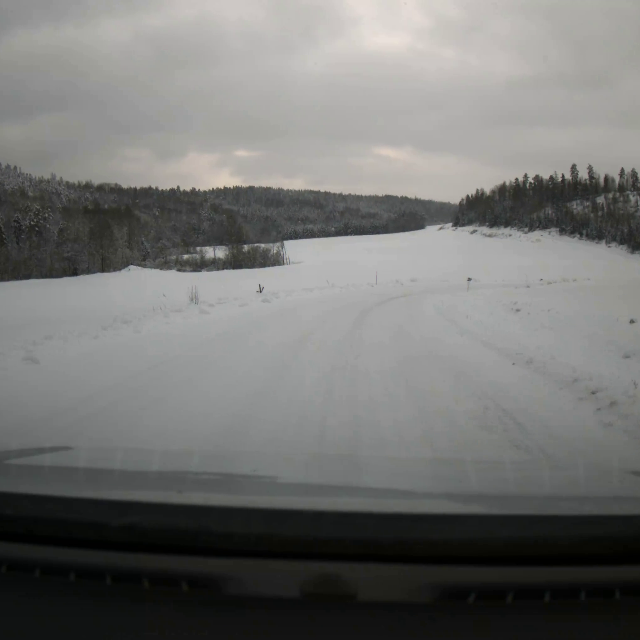} &
\vspace{0in}
\includegraphics[width = 1.15in]{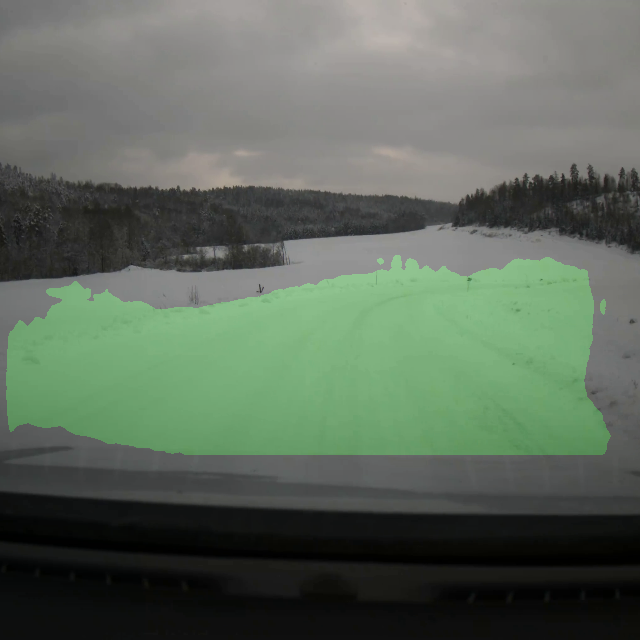} &
\vspace{0in}
\includegraphics[width = 1.15in]{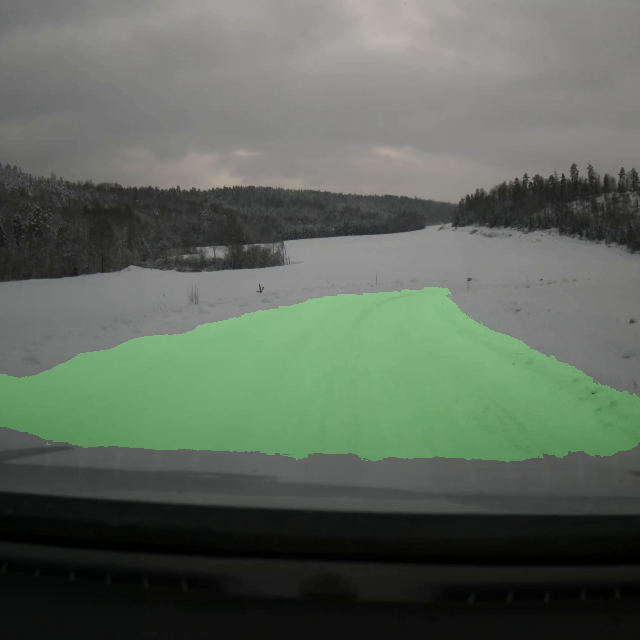}

\end{tabular}
\caption{A model trained with our proposed TADAP auto-labels demonstrating better detection of the drivable area than the previous state-of-the-art in an intersection where the drivable area has a complex shape and in a snowy countryside road that blends into the background.}
\label{intro}
\end{figure}

From these premises, a novel auto-labeling method combining trajectory-based learning with self-supervised features is presented, continuing from our previous work \cite{alamikkotervo2023position}. More precisely the method is performing Trajectory-Aided Drivable area Auto-labeling with Pre-trained self-supervised features (TADAP). In TADAP, a synchronized time series of images and Global Navigation Satellite System (GNSS) poses is first collected, and then the future GNSS poses for each frame are determined from the prerecorded data. A sample of the drivable area is extracted by projecting the future GNSS trajectory to the image frame. Then similarity with the sample area is determined using pre-trained self-supervised features. A feature vector is extracted for each image patch, and drivability is evaluated based on the similarity between each patch feature and the mean feature of the trajectory. The proposed method is validated with demanding winter driving data, but the use case can be expanded to any conditions. For further evaluation, a prediction model is trained with the TADAP labels, outperforming the self-supervised state-of-the-art of drivable area detection.  

The contributions of this paper are:
\begin{itemize}
    \item A novel method for automatically labeling drivable area in images based on feature similarity of pre-trained self-supervised features, outperforming the state-of-the-art (Figure \ref{intro}). 
    \item Utilizing GNSS for extracting the driven area. In previous work, 3D lidar-based point cloud mapping has been used \cite{seo2023learning,seo2023scate,schmid2022self}. 
    \item Validation with a dataset focused only on demanding winter driving. In the previous work, only a small part of the data has been recorded in winter conditions \cite{seo2023learning}. 
\end{itemize}

\section{Related work}
Robust detection of the drivable area in all conditions is required for autonomous driving and advanced driver assistance systems, but research has been mostly focused on favorable weather. Developing models for adverse conditions is challenging because the available data is limited, especially for winter driving. Recently, datasets that include winter data have been presented but the number of frames with drivable area labels is small, or drivable area labels are not provided at all. Thus, supervised learning approaches are difficult to implement in winter driving conditions. To remove the need for labeled data and provide better adaptability to changing conditions, self-supervised approaches have been proposed, but the performance is not yet comparable to supervised models. 

\subsection{Classic and supervised approaches}
In favorable weather and road conditions, the drivable area can be detected based on the color, texture, or lane markings \cite{chen1997vision,chiu2005lane,graovac2012detection}. However, especially in winter driving conditions, lane markings can be missing and the color and texture of non-drivable areas, like sidewalks, can be similar to the drivable area, making these approaches infeasible to use in most real-life applications. 

Machine learning approaches enable learning more advanced detection mechanisms from data. Currently, supervised deep-learning methods offer the best performance. The state-of-the-art models for supervised semantic traffic scene understanding not only predict the drivable area, but also other classes like pedestrian, car, and sidewalk \cite{tao2020hierarchical,borse2021inverseform,wang2023internimage}. These models perform well on conditions that are present in the training data but even a small domain shift between the training and test case compromises the performance. Cityscapes \cite{cordts2016cityscapes} and KITTI \cite{fritsch2013new} are both collected in favorable weather conditions but there is a significant drop in performance when training with one and testing with the other \cite{bolte2019unsupervised}. 

Training supervised methods for drivable area detection in winter conditions is challenging because the available data is very limited. Out of all available winter driving datasets only three \cite{sakaridis2021acdc,diaz2022ithaca365,kurup2023winter} provide drivable area labels (Table \ref{datasets}), and even then the number of labeled frames is quite low. 

\begin{table}[h]
    \centering
    \caption{Public datasets that include data collected in winter driving conditions}
    \begin{tabular}{lcc}
        & Number of frames with & \\
        Dataset & snow and drivable area label & Label type \\
        \hline
        CADCD \cite{pitropov2021canadian} & $\times$ & $\times$ \\
        ADCD \cite{sakaridis2021acdc} & 1k & semantic (image) \\
        Ithaca365 \cite{diaz2022ithaca365} & 3k & amodal (image)  \\
        WADS \cite{kurup2023winter} & 1k & semantic (3D-lidar scan) \\
        RADIATE \cite{sheeny2021radiate} & $\times$ & $\times$  \\
        DENSE \cite{bijelic2020seeing} & $\times$ & $\times$ \\
        \hline
    \end{tabular}
    \label{datasets}
\end{table}

WADS \cite{kurup2023winter} includes semantically labeled lidar scans that include the road class. Point cloud labels can be projected to the image frame quite accurately, but false positives can be caused by areas that are labeled in the point cloud but occluded in the image. ADCD \cite{sakaridis2021acdc} includes labels in the image frame, but the drivable area is mostly free of snow. Ithaca365 \cite{diaz2022ithaca365} provides amodal labels that include also occluded areas not visible in the image frame, which is not desired here. 

Despite the lack of data, there have been efforts to develop drivable area detection in winter conditions with supervised learning \cite{vachmanus2020semantic,lei2020semantic,diaz2022ithaca365,sakaridis2021acdc,rawashdeh2023camera}. However, the training distribution is narrow, while winter driving conditions have high variation because the road network is very extensive and snow can cause the appearance of the drivable area to change rapidly. For these reasons, we don't find supervised learning a viable option for general drivable area detection in winter conditions. 

\subsection{Self-supervised approaches}
Adaptability can be improved by using self-supervised approaches that learn to detect the drivable area without human supervision. Drivable area labels can be automatically generated by measuring the flatness of the environment using a 3D sensor and defining all flat continuous areas as drivable. Commonly stereo camera \cite{wang2019self,mayr2018self,cho2018multi,shinzato2014road} or 3D-lidar \cite{liu2018co,chen2023learning,shinzato2014roadlidar} is used. However, these methods struggle to reject regions that have similar height characteristics to the drivable area, such as sidewalks. Also, these methods have been validated with driving data from favorable weather conditions and some additional challenges would likely emerge in winter driving conditions as the drivable area can be uneven due to snow.  

Drivable area detection in winter driving conditions and off-road driving conditions have many common features. In both scenarios, the drivable area appearance can have high variation, and the boundaries between drivable and non-drivable regions are unclear. Self-supervised drivable area detection approaches for off-road conditions commonly utilize the vehicle trajectory to learn which areas are drivable \cite{schmid2022self,seo2023learning,zurn2020self,wellhausen2019should,seo2023scate, tang2017one,lieb2005adaptive}. However, only part of the drivable area is included in the trajectory, making it difficult to separate drivable areas not present in the data from non-drivable areas. 

The recent work of Seo et. al. \cite{seo2023learning} in this field utilizes self-supervised visual feature learning \cite{chen2020simple} to achieve better generalization to drivable areas not present in the trajectory. A pre-trained PSPNet \cite{zhao2017pyramid} is used as a starting point, but weights are not frozen. The model continues to learn the feature representations during the training process using contrastive loss. This approach yielded state-of-the-art results in multiple off-road and on-road scenes, including winter driving. 

\section{Research gap}
The quality of the self-supervised visual features used in the current trajectory-based approaches could be improved by increasing the backbone model size, using more training data, and a more advanced learning scheme. A wide variety of pre-trained models have been published in this scope \cite{caron2021emerging,zhou2021ibot,oquab2023dinov2,ilharco_gabriel_2021_5143773,he2022masked,caron2020unsupervised}, providing meaningful feature representations, suitable for downstream tasks like segmentation without fine-tuning. The quality of these features has been demonstrated by unsupervised segmentation models, that learn the mapping from pre-trained self-supervised features to segmentation labels \cite{hamilton2022unsupervised,kim2023causal} yielding state-of-the-art performance in unsupervised segmentation. In this paper, the state-of-the-art of trajectory-based drivable area detection is improved by using pre-trained self-supervised vision features \cite{oquab2023dinov2} instead of trying to learn the feature representations ourselves. The scalability is also improved by utilizing only GNSS for trajectory extraction instead of 3D-lidar-based point cloud mapping \cite{seo2023learning,seo2023scate,schmid2022self}.

\section{Methods}

Our drivable area auto-labeling method, TADAP,  first saves a synchronized time series of images and GNSS poses and projects the future GNSS trajectory to the image based on the pre-collected data. The image area corresponding to the trajectory is used as a reference for detecting the whole drivable area. The likelihood of an image area belonging to the drivable area is estimated based on the similarity with the mean feature of the reference area. The similarity is evaluated using pre-trained self-supervised features and the output is refined using Conditional Random Field (CRF) post-processing \cite{krahenbuhl2011efficient}, yielding drivable area labels with no human supervision. The TADAP-auto labeling includes the following steps:
\begin{itemize}
\item A) GNSS trajectory extraction (Figure \ref{TADAP} A)
\item B) Similarity with trajectory (Figure \ref{TADAP} B1-B3)
\item C) CRF post-processing (Figure \ref{TADAP} C)
\item D) Second iteration update (Figure 3 D)
\end{itemize}
All steps are presented in detail below. The data used for validation is described in Section \ref{sec:data} and the validation implementations are described in Section \ref{sec:validation}.\\

\begin{figure*}[h]
    \centering
    \includegraphics[width=\textwidth]{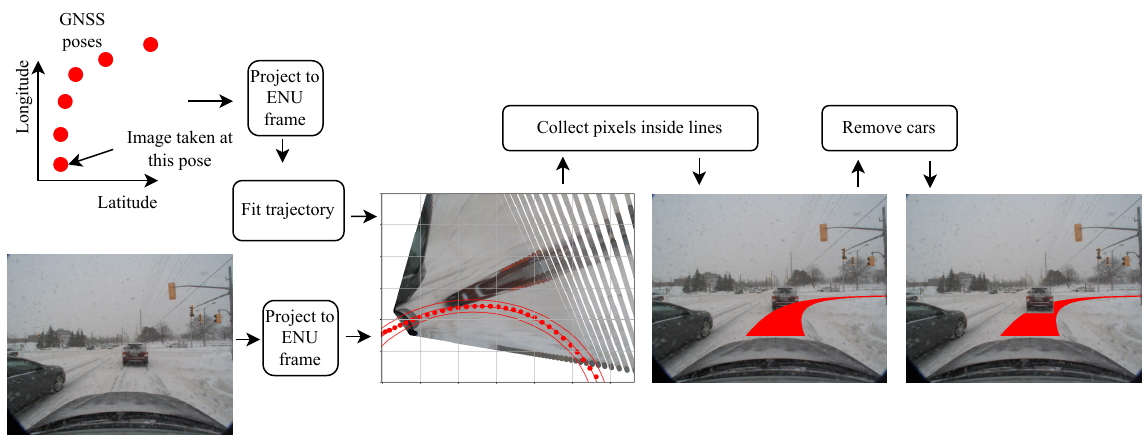}
    \caption{GNSS trajectory extraction. The image is projected to the ENU frame with a planarity assumption and then the pixels inside the fitted circles are collected. GNSS poses for the following 50 m of driving are considered for each image.} 
    \label{mask_extraction}
\end{figure*}

\subsection{GNSS trajectory extraction}
In the previous work, the future trajectory is collected and projected to the image frame based on 3D point cloud mapping of the route\cite{seo2023learning,seo2023scate,schmid2022self}. This approach has high accuracy even far ahead of the vehicle. However, the mapping process can be time-consuming and requires an expensive 3D-lidar sensor, making the method hard to scale. In this paper, a GNSS-based method for extracting the image areas corresponding to the trajectory is presented (Figure \ref{mask_extraction}). 

In our approach, for each image, a series of GNSS poses corresponding to the following 50 m of driving on the data collection route is considered. The image and the GNSS poses are projected to the same East-North-Up (ENU) frame. ENU is a local tangent plane Cartesian coordinate frame mounted to a reference point on the surface of the earth. The positive x-axis points to the east, the positive y-axis to the north, and the positive z-axis up. ENU coordinate representation enables easy projection from the image plane to the ground plane and the error is small if the distance from the origin point is small.

The image was projected to the GNSS receiver frame with the assumption of the ground being planar. The homography transform \textbf{H} from the pixel coordinates $(u,v)$ to the ground plane coordinates in the GNSS receiver frame $(X,Y)$ is defined as: 

\begin{equation}
\label{homography transform}
\begin{aligned}
s
\begin{bmatrix}
X\\
Y\\
1\\
\end{bmatrix}
=
\underset{3\times3}{\boldsymbol{H}}
\begin{bmatrix}
u\\
v\\
1\\
\end{bmatrix},
\end{aligned}
\end{equation}
\\

where $s$ is the scaling factor. The homography transform $\boldsymbol{H}$ was determined by measuring four point pairs $[(u,v),(X, Y)]$ and finding the corresponding transform with the OpenCV \cite{bradski2000opencv} findHomography method. 

Finally, the pixels were transformed from the GNSS receiver frame to the ENU frame, yielding the image pixels and GNSS poses in the same ENU frame. A circle was fitted to the GNSS poses with the least square error criteria using Scipy optimize package \cite{2020SciPy-NMeth} and then all pixels closer than half of the vehicle's width from the fitted arc were collected. GNSS poses for the following 50 m of driving were considered for each image. A circle provides a good fit for poses both in straight and curved segments.

In some cases, there were other vehicles in the future trajectory. They had to be removed from the trajectory mask because it is not desirable to include vehicles in the drivable area sample. In previous work, occluded areas have been removed from the trajectory mask based on 3D-point cloud mapping  \cite{seo2023learning,schmid2022self}. In our approach, point cloud mapping is not available so vehicles were detected in images using pre-trained Yolo v5 \cite{glenn_jocher_2020_4154370} and the pixels inside the bounding box were removed from the trajectory mask. This approach was found to be reliable, even with multiple vehicles in the frame.  

\subsection{Similarity with trajectory}
Our TADAP labeling process estimates drivability by computing the similarity between each image patch feature and the mean feature of the trajectory. The similarity computation includes DINOv2 feature extraction, trajectory mean feature computation, and feature similarity computation. 

\subsubsection{DINOv2 feature extraction}
In this paper, the objective was to generalize the drivable area detection to areas not present in the trajectory by using a self-supervised feature extractor (Figure \ref{TADAP} B1) pre-trained with a large dataset. Here a pre-trained DINOv2 \cite{oquab2023dinov2} was used with the parameters specified in Table \ref{DINOparams} based on its state-of-the-art performance in dense feature extraction. 

\begin{table}[H]
    \centering
    \caption{Feature extraction parameters}
    \begin{tabular}{l|l}
        Backbone & DINOv2 ViT-gigantic (1.1 B params) \\
        Patch size & 14x14 pixels \\
        Patch feature vector size & 1536 \\
        Image input resolution & 644x644 \\
        Feature space resolution & 46x46 (644/14) \\
    \end{tabular}
    \label{DINOparams}
\end{table}

\noindent  DINOv2 uses a ViT backbone \cite{dosovitskiy2020image} that divides the input image into patches of 14x14 pixels. Each patch has a feature vector of length 1536 that represents the contents of that patch. The image patch features are extracted after the last transformer block of the model to yield the most refined representation. The DINOv2 backbone has been pre-trained with a large dataset using a self-supervised approach resulting in a deep understanding of image representations. As a result, image patches with similar contents have similar feature vectors.  

\subsubsection{Trajectory mean feature computation}
 
Computing the correspondence between a single patch from the trajectory and each image patch would provide a simple solution for determining similarity with the trajectory. However, in this paper similarity with each image area and the trajectory mean feature is computed to produce more reliable results. The mean feature provides a more accurate reference point that represents the center of the trajectory feature cluster. The mean feature $f_ {mean}$ computation (Figure \ref{TADAP} B2) is defined as:

\begin{equation}
\label{attention}
\begin{aligned}
\boldsymbol{f}_{mean}=\frac{1}{N}\sum_{\substack{patch\in\\ trajectory}} \boldsymbol{f}_{patch},
\end{aligned}
\end{equation}

where $f_{patch}$ is the patch feature vector and $N$ is the number of patches in the trajectory. 

\subsubsection{Feature similarity computation}

The similarity with the mean feature (Figure \ref{TADAP} B3) is evaluated based on cosine similarity. Cosine similarity is usually preferred over Euclidian distance when comparing high-dimensional features and it is commonly used when comparing similarity of self-supervised vision features, including the Dinov2 features that have been used here \cite{oquab2023dinov2}. Cosine similarity is the cosine of the angle between two vectors and can be defined as:

\begin{equation}
\label{attention}
\begin{aligned}
cos(\theta)=\frac{\boldsymbol{f}_a \cdot \boldsymbol{f}_b}{\|\boldsymbol{f}_{a}\| \: \|\boldsymbol{f}_{b}\|},
\end{aligned}
\end{equation}

\noindent where $\boldsymbol{f}_a$ and $\boldsymbol{f}_b$ are vectors, $\theta$ the angle between the vectors and $\|\|$ is euclidian norm. Now the similarity between each patch feature vector and the trajectory mean feature can be computed from:

\begin{equation}
\label{attention}
\begin{aligned}
F_{patch}=\frac{\boldsymbol{f}_{patch} \cdot \boldsymbol{f}_{mean}}{\|\boldsymbol{f}_{patch}\| \: \|\boldsymbol{f}_{mean}\|},
\end{aligned}
\end{equation}

The similarities are finally normalized by dividing with the maximum similarity $F_{max}$ inside the frame: 

\begin{equation}
\label{norm_sim}
\begin{aligned}
F_{patch}^{norm}=\frac{F_{patch}}{F_{max}}.
\end{aligned}
\end{equation}

In our TADAP auto-labeling method the normalized feature similarity $F_{patch}^{norm}$ with the trajectory mean feature describes the likelihood that a given image patch is part of the drivable area.  

\subsection{CRF post-processing}
\label{crf}
The normalized similarity (\ref{norm_sim}) between an image patch and the trajectory mean feature describes the likelihood that it is part of the drivable area. As the spatial resolution of the patches is low, CRF post-processing (Figure \ref{TADAP} C) is used to refine the labels. For comparison, results are also presented without any post-processing, using 0.5 as a hard threshold for the required similarity score. Here, CRF post-processing is implemented with pyndensecrf library \cite{krahenbuhl2011efficient} similarly to \cite{hamilton2022unsupervised}. Here, parameters $a=4$, $b=3$, $\theta_{\alpha}=25$, $\theta_{\beta}=3$ and $\theta_{\gamma}=5$ with 10 iterations provided good results based on visual evaluation. When auto-labeling, the unary potentials are initialized with the negative logarithm of the normalized similarity (\ref{norm_sim}), and when predicting, with the negative logarithm of the projection head output. The unary potentials represent the confidence of assigning a specific pixel to the drivable area class.

\subsection{Second iteration update}
\label{2_iter_update}
\begin{figure*}[t]
    \centering
    \includegraphics[width=\textwidth]{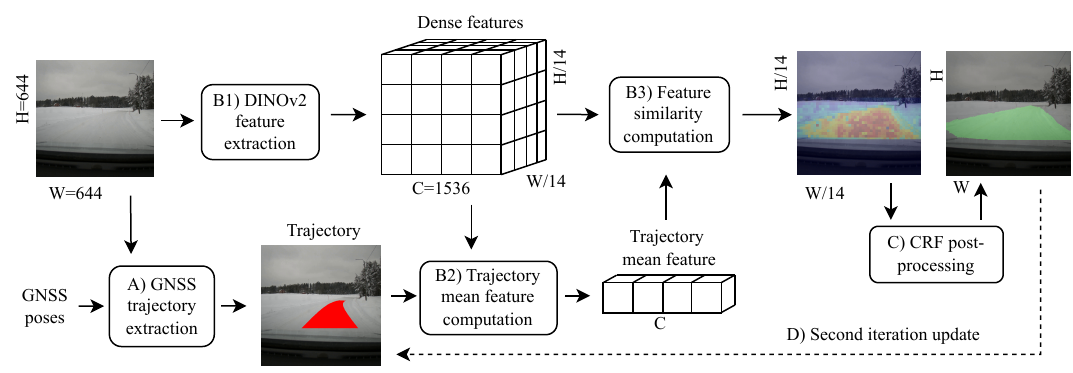}
    \caption{TADAP-autolabeling process. A) The trajectory is extracted based on GNSS. B1) Features are extracted for each image patch using a pre-trained self-supervised feature extractor. B2) The mean feature of the trajectory is determined. B3) Similarity with the mean feature is computed. C) Label resolution is refined with CRF post-processing. D) The process from B2 to C is repeated using the auto-label from the first iteration as the trajectory sample. All steps are described in detail in the corresponding sections.}
    \label{TADAP}
\end{figure*}

The normalized similarity (\ref{norm_sim}) is already a good metric for detecting the drivable area from the image, but as the trajectory mostly includes patches from the vehicle's own lane, the mean feature is biased. The bias can be decreased by repeating the similarity computation twice. In the first iteration similarity with the trajectory mean feature is computed yielding a drivable area prediction that is a significantly better estimate of the whole drivable area than the trajectory itself, but it struggles to detect areas far away from the trajectory, like adjacent lanes. Then the mean feature of the drivable area prediction is computed (Figure \ref{TADAP} D) to obtain a less biased reference point of the whole drivable area. Finally, similarity with the improved mean feature is computed, yielding a better estimate of the whole drivable area. Repeating the process more than two times yielded no additional improvements in our experiments.  

\subsection{Data}\label{sec:data}
A dataset with synchronized GNSS pose-image pairs and calibration parameters for projection from the ground plane to the image frame is required for generating TADAP labels. Additionally, a manually labeled test set is needed for testing. As no suitable dataset was available, a new winter driving dataset was collected. The dataset was collected in Espoo  Finland, and its proximity on two different days in February 2023. The data includes four scenes: suburban, highway, countryside, and intersection. In the countryside scene, the drivable area is fully covered in snow, in the highway scene the drivable area is only partially covered in snow, and in the suburban scene, the snow cover varies from partial to full. The intersection scene includes intersection cases from all the previous scenes. It was separated into its own scene because the drivable area shape in intersections is more complex and the auto-labeling process needs to extend its detection further from the trajectory to capture the whole drivable area. 

The data collection platform was a 2019 Ford Focus retrofitted with a high-quality sensor setup and an onboard computer. The sensors used were:
\begin{itemize}
    \item Novatel PWRPAK 7DE2 GNSS/INS unit
    \item FLIR blackfly S 2448x2048 forward-facing machine vision camera 
\end{itemize}

The raw data contained 184k frames captured with 20Hz frequency. From the raw data, every fifth frame was retained leaving 37k frames. The GNSS translations were extracted from the Novatel/oem7/bestpos topic which was published at 10 Hz frequency and the heading angle was extracted from the Novatel/oem7/inspva topic which was published at 50 Hz frequency. The Novatel measurement unit fuses an Inertial Navigation System (INS) with GNSS for more accurate pose estimates. Time synchronization was achieved by pairing images with the GNSS position with the closest timestamp. 

1000 random samples from the data set were labeled and divided into 20-80 test-validation split that was kept unchanged for all experiments. The scene distribution of the labeled images is suburb: 43 \%, highway: 19 \%, countryside: 19 \% and intersection: 19 \%. All labeled frames were removed from the training data as well as 10 previous frames and 10 following frames for each of them. This approach maximizes the variance in the training, validation, and test data while minimizing the similarity between the training set and the validation and test set. The final composition of the dataset is 18k training images, 0.2k validation images, and 0.8k test images. Only validation and test images have manual labels.  

\subsection{Validation}\label{sec:validation}
The TADAP labels can be used to train any segmentation model. Here, a simple linear projection head is trained on top of the Dinov2 feature extractor to demonstrate the performance achieved when training with the TADAP labels. For increased resolution, CRF post-processing is implemented as described in Section \ref{crf}. Different learning rates from 1e-3 to 1e-7 with tenfold increments between experiments were tested. A learning rate of 1e-4 yielded the best performance when training the projection head with a batch size of 64. The projection head was trained for 50 epochs and the checkpoint with the highest validation score was chosen for the test.

The model trained with the TADAP labels was benchmarked against the state-of-the-art of self-supervised drivable area estimation by Seo et. al. \cite{seo2023learning}. The model was trained with an input size of 640x640, instead of 644x644 used with the TADAP labels as the input size had to be divisible by eight. The maximum achievable batch size of 16 was used when training with two Nvidia A100 GPUs in parallel. In the original publication learning rate of 1e-3 was used, but with our setup, it caused the model to quickly overfit only to the trajectory. To achieve optimal performance, different learning rates were tested from 1e-3 to 1e-7 with tenfold increments between tests. A learning rate of 1e-6 was found to yield the best performance here. The model was trained for 60 epochs and after each epoch, different detection thresholds from 0.1 to 0.9 with 0.1 increments were tested with the validation set to find the best-performing model checkpoint and the corresponding detection threshold. The optimal checkpoint for CRF post-processing was chosen by applying CRF post-processing as described in Section \ref{crf} after each epoch before running validation and choosing the checkpoint with the highest validations score. 

For both methods, a single model was trained with all of the data, and test results were presented for each scene with this model to mimic a real-life scenario where a single model needs to adapt to different driving conditions. Intersection over Union (IoU), F1 score (F1), precision (PRE), and recall (REC) were used as evaluation metrics.

In evaluation, regions above the horizon and below the hood of the car were not considered for any of the methods. The horizon limit was placed at a conservatively high fixed position at 240 pixels from the top for a 640x640 image size. This approach mistakenly excluded some drivable areas in steep uphill or downhill scenarios but it was very rare. 

\section{Results}
\begin{figure*}[h]
\centering
\setlength\extrarowheight{-4pt}
\begin{tabular}{@{}c@{\hspace{2pt}}c@{\hspace{4pt}}c@{\hspace{4pt}}c@{\hspace{4pt}}c@{}}
&
&
&
&
\footnotesize Model trained with\\

&
\footnotesize Input &
\footnotesize Seo et. al. &
\footnotesize TADAP labels &
\footnotesize TADAP labels \\

\vspace{3pt}

\parbox[t]{2mm}{\rotatebox[origin=c]{90}{\footnotesize Suburb}} &
\includegraphics[width = 1.5in,valign=m]{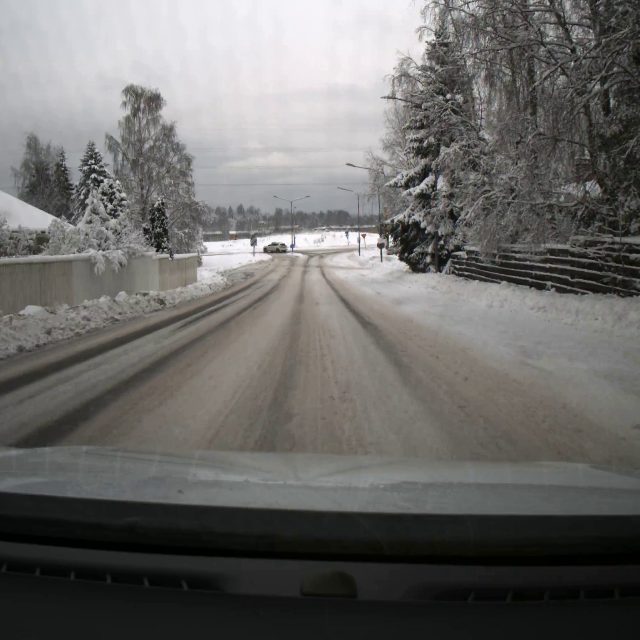} &
\includegraphics[width = 1.5in,valign=m]{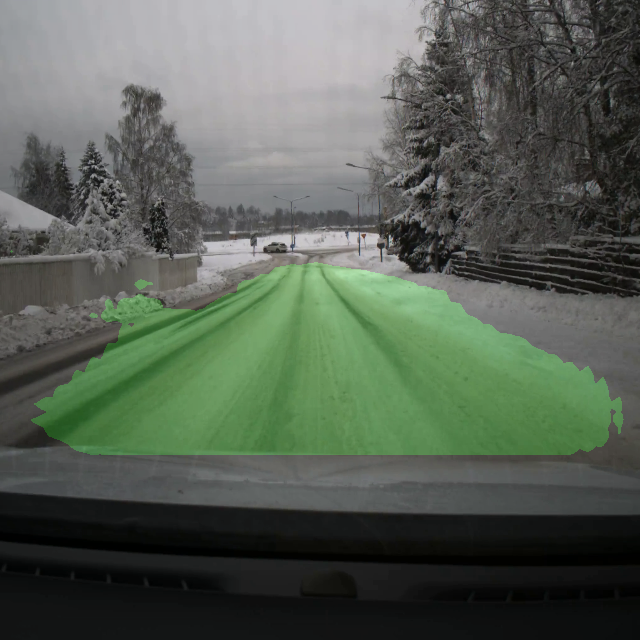} &
\includegraphics[width = 1.5in,valign=m]{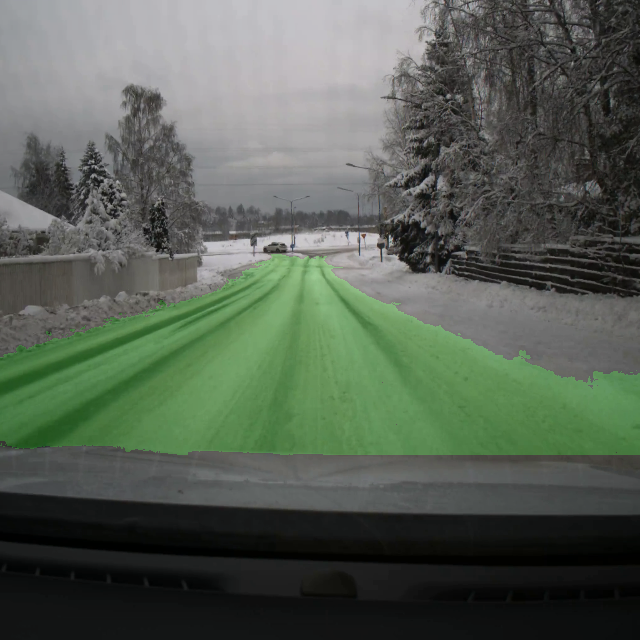} &
\includegraphics[width = 1.5in,valign=m]{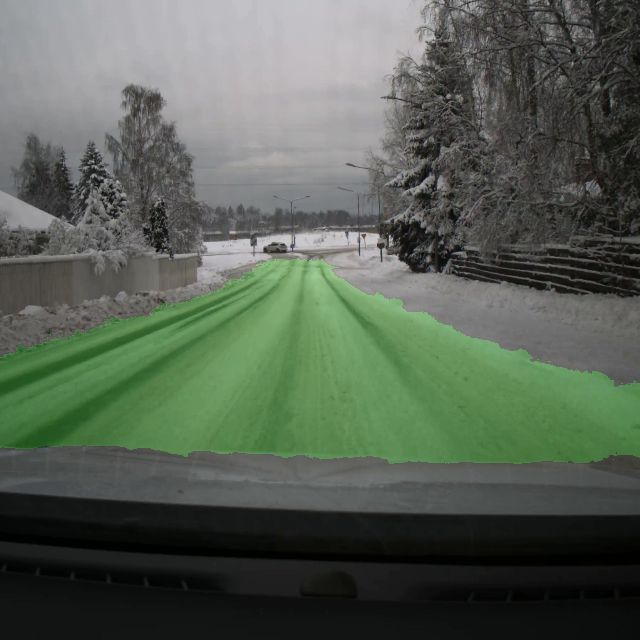} \\

\vspace{3pt}

\parbox[t]{2mm}{{\rotatebox[origin=c]{90}{\footnotesize Highway}}} &
\includegraphics[width = 1.5in,valign=m]{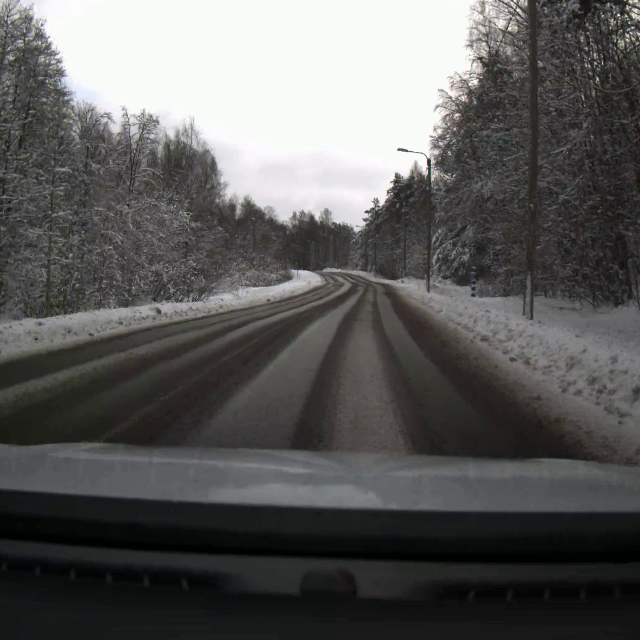} &
\includegraphics[width = 1.5in,valign=m]{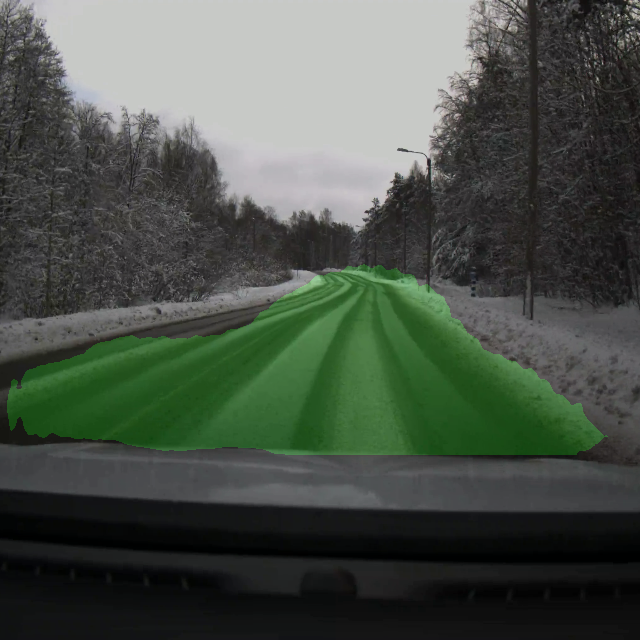} &
\includegraphics[width = 1.5in,valign=m]{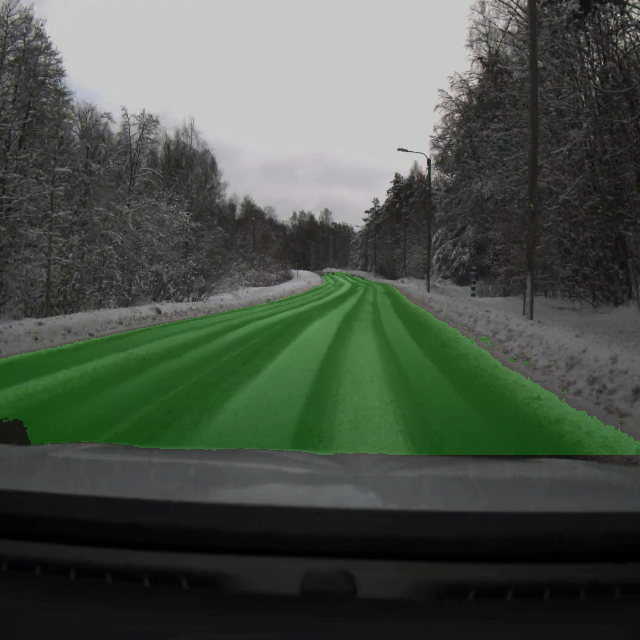} &
\includegraphics[width = 1.5in,valign=m]{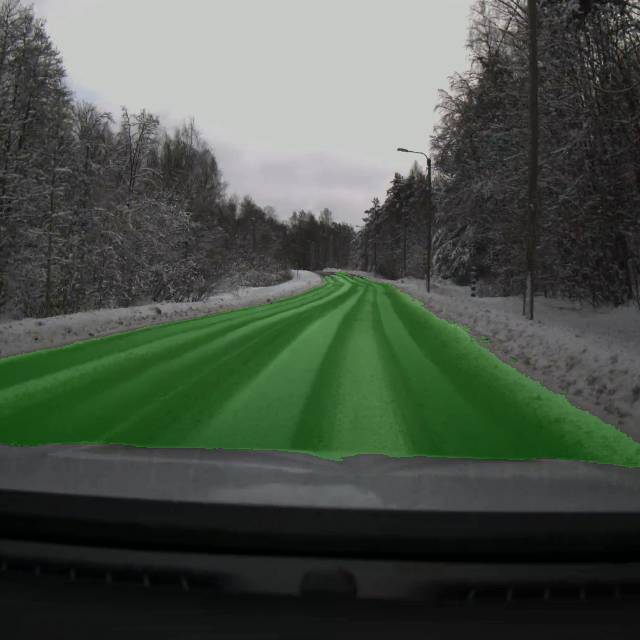} \\

\vspace{3pt}

\parbox[t]{2mm}{{\rotatebox[origin=c]{90}{\footnotesize Countryside}}} &
\includegraphics[width = 1.5in,valign=m]{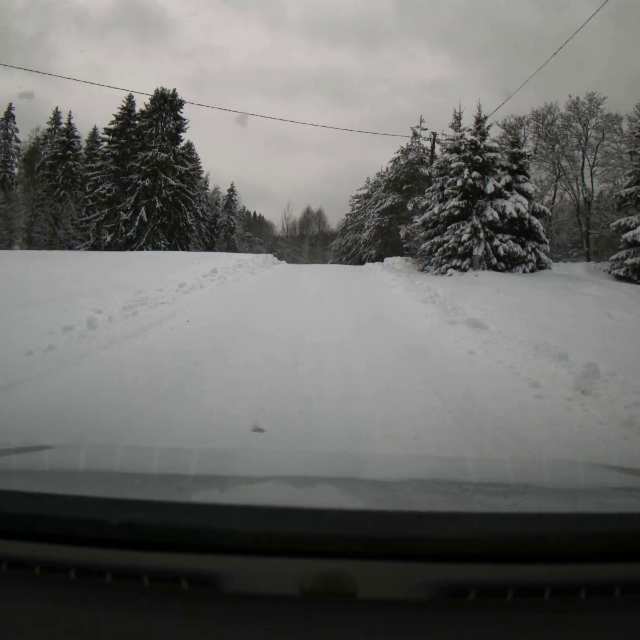} &
\includegraphics[width = 1.5in,valign=m]{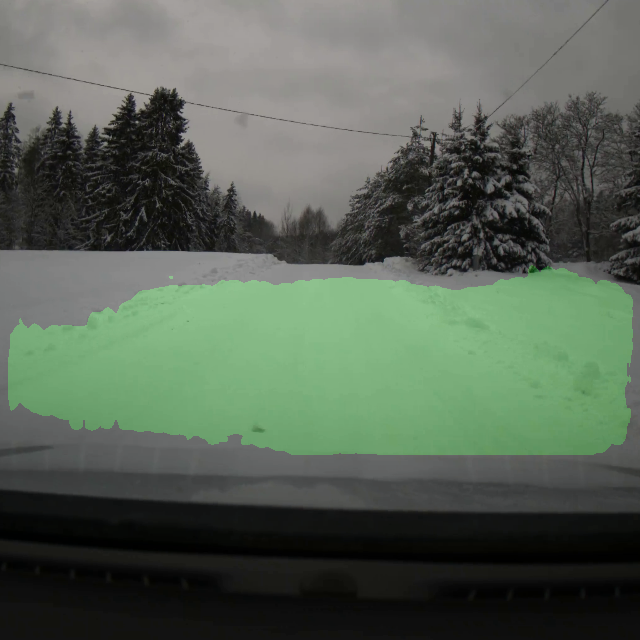} &
\includegraphics[width = 1.5in,valign=m]{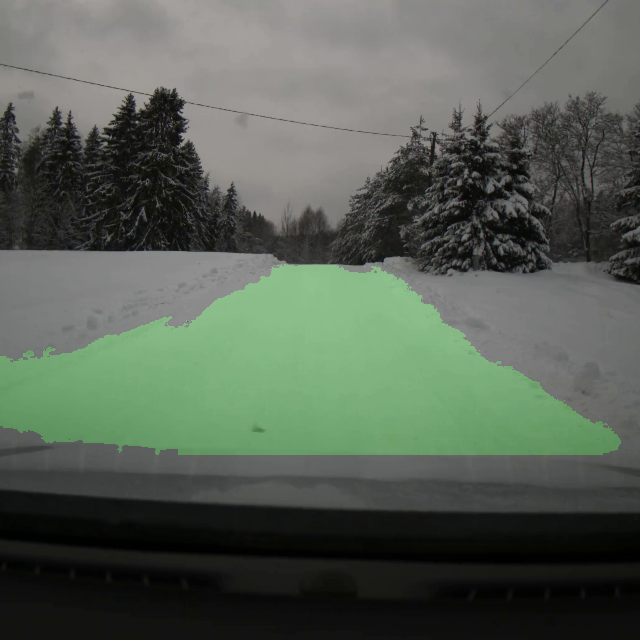} &
\includegraphics[width = 1.5in,valign=m]{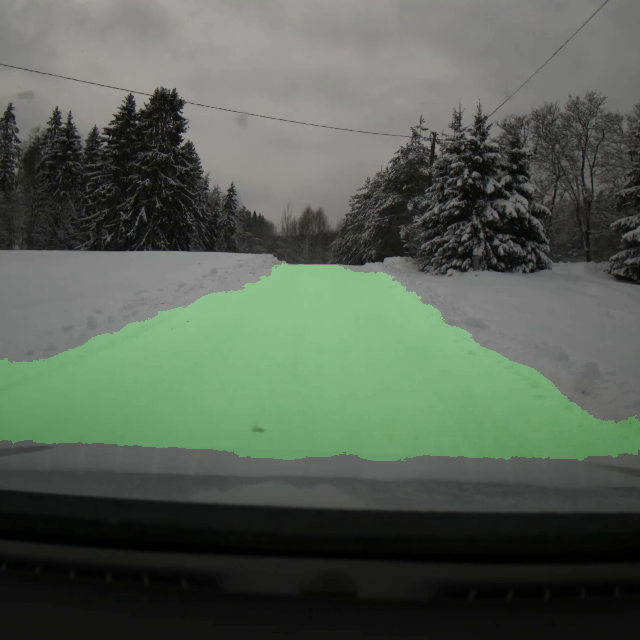} \\

\vspace{3pt}

\parbox[t]{2mm}{{\rotatebox[origin=c]{90}{\footnotesize Intersection}}} &
\includegraphics[width = 1.5in,valign=m]{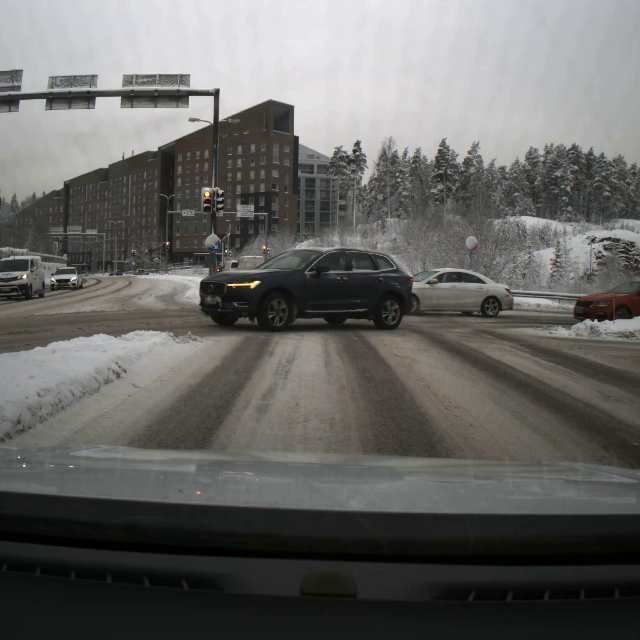} &
\includegraphics[width = 1.5in,valign=m]{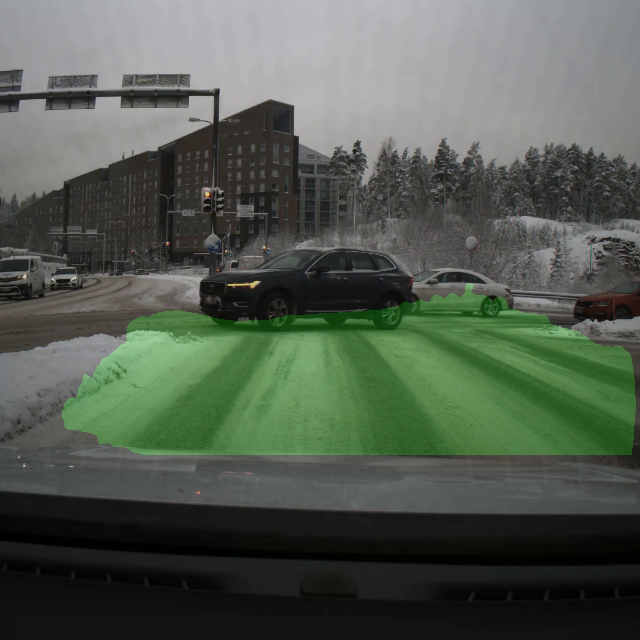} &
\includegraphics[width = 1.5in,valign=m]{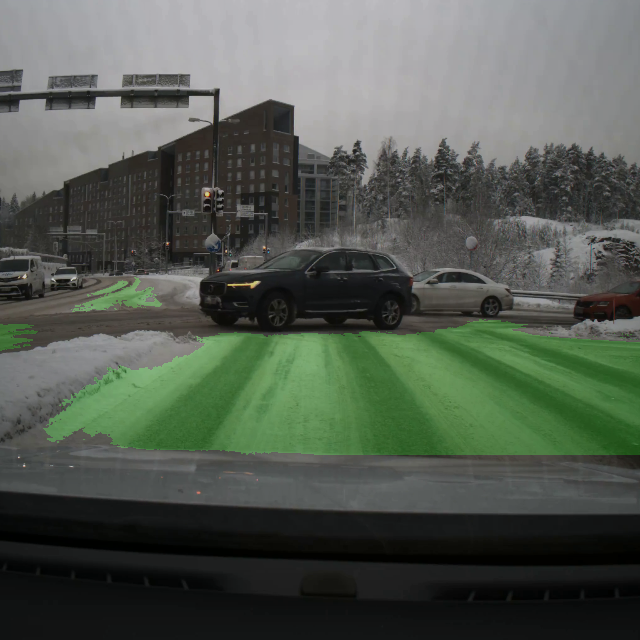} &
\includegraphics[width = 1.5in,valign=m]{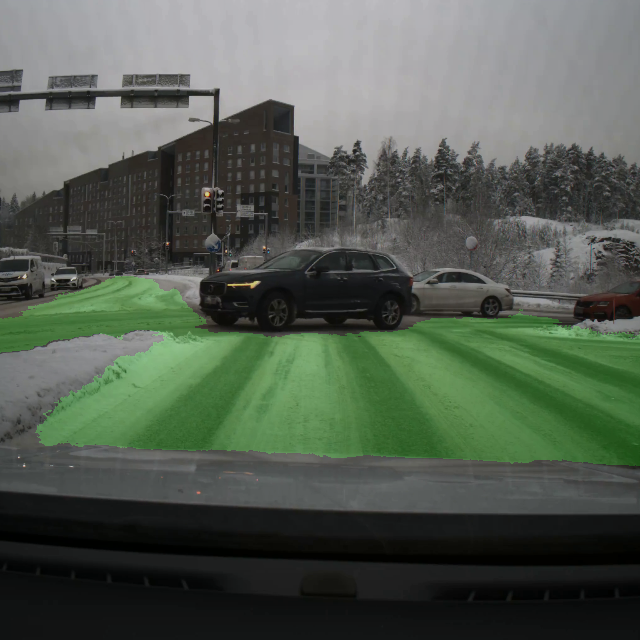}\\

\end{tabular}
\caption{Samples of predictions from the test set. For each method, the best-performing configuration is used: Seo et. al. without CRF post-processing, TADAP labels with second iteration update, and CRF post-processing and the model trained with TADAP labels with CRF post-processing.}
\label{qualitative results}
\end{figure*}

In this section, the performance of the TADAP labels is evaluated (Table \ref{tab:labels}) and the quality of the labels is further tested by training a prediction model using the TADAP labels. The performance of this model is compared against the state-of-the-art of self-supervised drivable area estimation (Table \ref{tab:prediction}). Samples of TADAP labels and predictions of the benchmarked models are presented in Figure \ref{qualitative results}. Precision and recall are presented in Table \ref{tab:pre_rec} and Figure \ref{fig:pre_rec}.

\subsection{TADAP labels}
\noindent For TADAP auto-labeling three different configurations were tested (Table \ref{tab:labels}). \textbf{Baseline} defines the label based on feature similarity with the trajectory mean feature. CRF post-processing or second iteration update is not used. \textbf{Second iteration update} repeats the feature similarity computation a second time using the first iteration estimate for determining the mean feature in the second iteration as described in Section \ref{2_iter_update}. Finally, \textbf{CRF post-processing} as described in Section \ref{crf} is added to the second iteration update. 

The advantages of the second iteration update and CRF post-processing were clear in all metrics and scenes. The second iteration update yielded a \textbf{+5.5}, and CRF post-processing an additional \textbf{+3.0} improvement in IoU. Including both of these steps improved recall by \textbf{+13.9} while precision decreased only by \textbf{-4.9} (Table \ref{tab:pre_rec}).     

\subsection{Prediction}
When predicting, the future GNSS trajectories are not available anymore. The prediction performance for a model trained with our TADAP labels was evaluated against the state-of-the-art of drivable area estimation by Seo et. al. For a fair comparison, results are presented with and without CRF post-processing for both methods (Table \ref{tab:prediction}). 

The model by \textbf{Seo et. al.} was trained with our dataset as described in the original publication. \textbf{CRF post-processing} was not used in the original publication but was applied here for a fair comparison. \textbf{Model trained with TADAP labels} trains a linear segmentation head on top Dinov2 backbone using the TADAP labels. \textbf{CRF post-processing} refines the prediction resolution. Implementation details are presented in Section \ref{sec:validation}. 

\begin{figure}[t]
    \centering
    \includegraphics[width=0.5\textwidth]{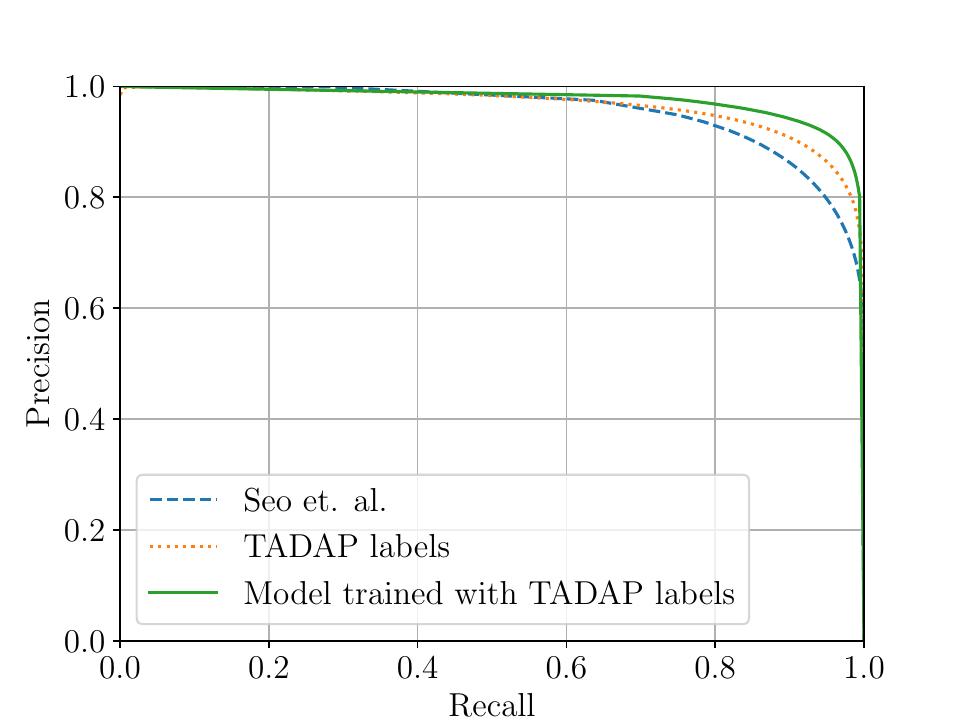}
    \caption{Precision-Recall curves for the previous state-of-the-art (Seo et. al.), TADAP labels with second iteration update, and a model trained with the TADAP labels. CRF post-processing output is binary so it is not used here with any of the methods.}
    \label{fig:pre_rec}
\end{figure}

The model trained with TADAP labels outperformed the previous state-of-the-art in all metrics and in all scenes. Our model achieved a \textbf{+9.6} improvement in IoU (Table \ref{tab:prediction}), a \textbf{+6.2} improvement in recall, and a \textbf{+4.9} improvement in precision (Table \ref{tab:pre_rec}). Our model detected more of the drivable area while mistaking less non-drivable areas as drivable at the same time. The precision-recall curve of our model was above the previous state-of-the-art and the TADAP labels it was trained with at all thresholds (Figure \ref{fig:pre_rec}). 

CRF post-processing didn't improve the performance of the previous state-of-the-art, while it increased the IoU achieved by our model by \textbf{+1.1}. This indicates that the output distribution of our model was closer to reality, as the performance of the CRF post-processing depends on the accuracy of the initial likelihoods provided.  

\begin{table*}[h]
\setlength{\tabcolsep}{15pt}
\centering
\caption{Ablation study for the TADAP labels. IoU and F1 score are reported for each scene.}
\begin{tabular}{l c c c c c}
& all & suburban & highway & countryside & intersection \\
& IoU \ F1 & IoU \ F1 & IoU \ F1 & IoU \ F1 & IoU \ F1 \\
\hline
\rowcolor{gray!10}
Baseline TADAP labels & 77.2 \ 86.9  & 77.4 \ 87.1 & 74.9 \ 85.3 & 80.1 \ 88.9 & 76.1 \ 86.3 \\
\rowcolor{gray!10}
+Second iteration update & 82.7 \ 90.3 & 80.2 \ 88.8 & 85.3 \ 92.0 & 84.8 \ 91.7 & 83.6 \ 90.9 \\
\rowcolor{gray!10}
+CRF post-processing=TADAP & 85.7 \ 92.1 & 83.5 \ 90.7 & 89.7 \ 94.5 & 86.1 \ 92.5 & 86.4 \ 92.5 \\
\hline
\end{tabular}
\label{tab:labels}
\end{table*}

\begin{table*}[h]
\setlength{\tabcolsep}{15pt}
\centering
\caption{Prediction benchmark for a model trained with the TADAP labels. IoU and F1 score are reported for each scene.}
\begin{tabular}{l c c c c c}
 & all & suburban & highway & countryside & intersection \\
& IoU \ F1 & IoU \ F1 & IoU \ F1 & IoU \ F1 & IoU \ F1 \\
\hline
Seo et. al. & 78.5 \ 87.7  & 77.2 \ 86.9 & 79.8 \ 88.6 & 79.7 \ 88.6 & 79.1 \ 87.9 \\
+CRF post-processing & 77.8 \ 87.2  & 76.5 \ 86.4 & 80.8 \ 89.3 & 74.8 \ 85.3 & 80.4 \ 88.8 \\
\rowcolor{gray!30}
Model trained with TADAP labels & 87.0 \ 92.9 & 84.6 \ 91.4 & 90.6 \ 95.0 & 88.2 \ 93.7 & 87.6 \ 93.3 \\
\rowcolor{gray!30}
+CRF post-processing & \textbf{88.1 \ 93.5} & \textbf{85.7 \ 92.1} & \textbf{91.8 \ 95.7} & \textbf{89.3 \ 94.3} & \textbf{88.9 \ 94.0} \\
\hline
\end{tabular}
\label{tab:prediction}
\end{table*}

\clearpage

\setlength{\textfloatsep}{30pt}
\setlength{\floatsep}{10pt}

\begin{table}[t]
    \centering
    \caption{Precision and recall}
    \begin{tabular}{l c c}
         & PRE & REC \\
         \hline
         Seo et. al & 87.4 & 89.0 \\
         +CRF post-processing & 84.1 & 91.9 \\
         \rowcolor{gray!10}
         Baseline TADAP labels & \textbf{95.6} & 80.3 \\
         \rowcolor{gray!10}
         +Second iteration update & 90.2 & 91.1 \\
         \rowcolor{gray!10}
         +CRF post-processing=TADAP & 90.7 & 94.2 \\
         \rowcolor{gray!30}
         Model trained with TADAP labels & 92.0 & 94.2 \\
         \rowcolor{gray!30}
         +CRF post-processing & 92.3 & \textbf{95.2} \\
         \hline
    \end{tabular}

    \label{tab:pre_rec}
\end{table}

\section{Discussion}
\vspace{-170pt}

\begin{figure}[b]
\setlength\extrarowheight{-4pt}
\begin{tabular}{cc}
\centering

\includegraphics[width = 1.5in]{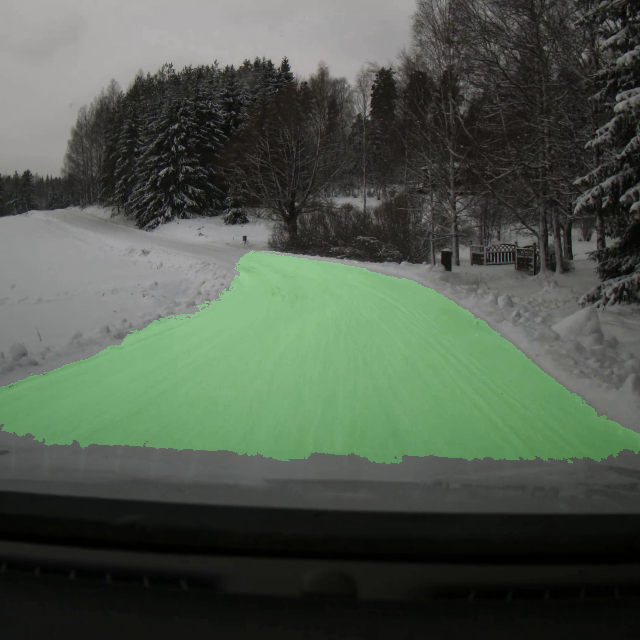} &
\includegraphics[width = 1.5in]{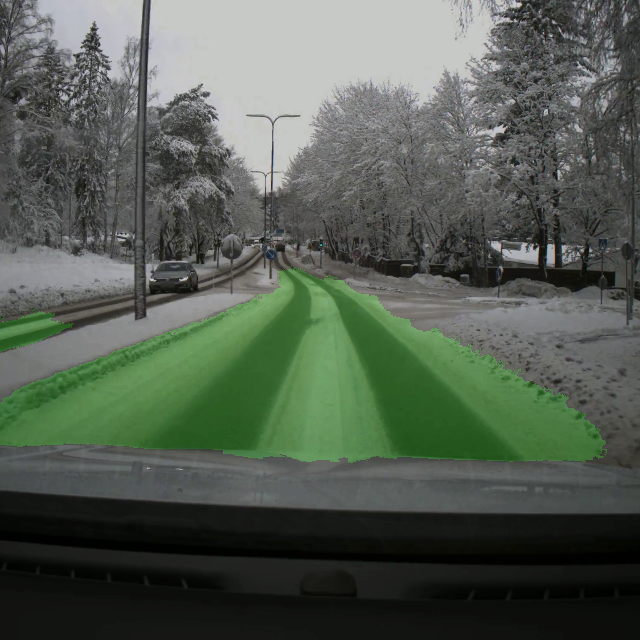} \\
\footnotesize a) drivable area far & \footnotesize b) oncoming lane \\
\footnotesize ahead not detected & \footnotesize not detected \\
\includegraphics[width = 1.5in]{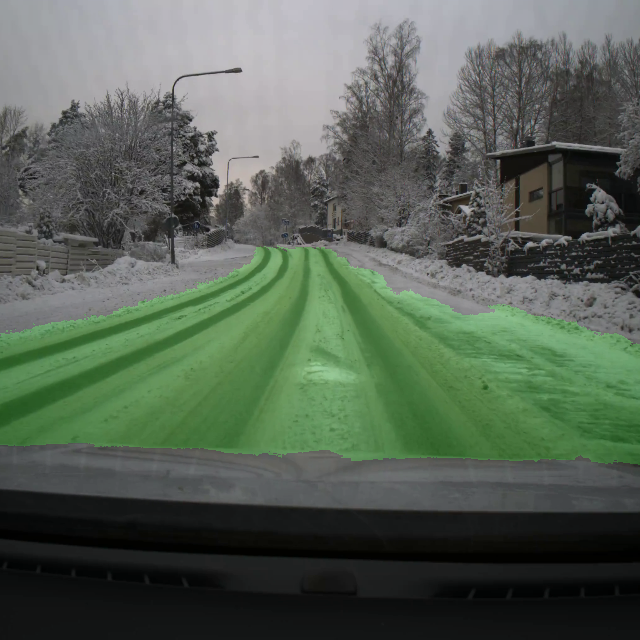} &
\includegraphics[width = 1.5in]{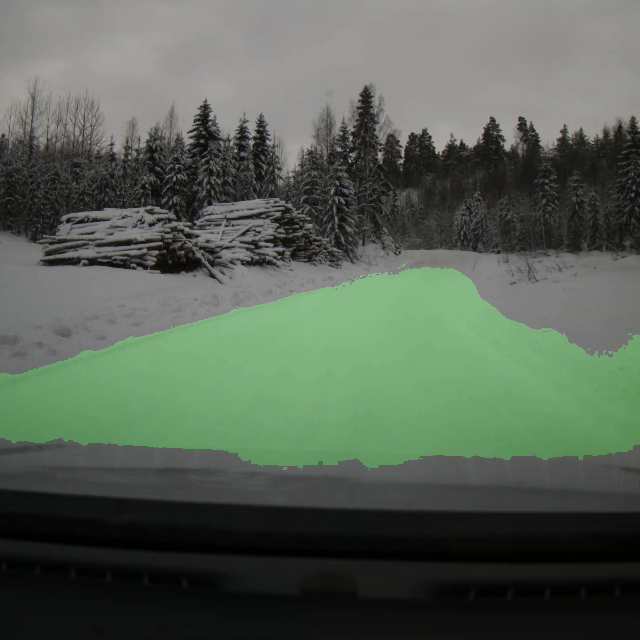}\\
\footnotesize c) sidewalk detected & \footnotesize d) road shoulder detected\\
\footnotesize as drivable & \footnotesize as drivable

\end{tabular}
\caption{Development areas of the TADAP labels.}
\label{struggle}
\end{figure}

The model trained with the TADAP labels outperformed the previous state-of-the-art of self-supervised drivable area detection in all driving scenes and all metrics. When training a model with the TADAP labels, the model reached higher performance than the TADAP labels it is learning from. When having access to labels from different scenarios the model can learn the difference between a drivable and non-drivable area more accurately than an individual label that only relies on the trajectory sample of that frame. Our versatile dataset provides a reliable evaluation of drivable area detection in winter conditions. In previous work, similarly sized datasets have been used, while only a portion of the data concentrated on winter driving.

\vspace{-28pt}

The TADAP labels have the highest performance in the highway scene and the lowest in the suburban scene. Interestingly, in the countryside scene, where the drivable area boundary is quite unnoticeable, and in the intersection scene, where the drivable area can have a complex shape, the performance is significantly higher than in the suburban scene. This behavior can be explained with the working principle of the TADAP labels. They rely fully on comparing similarity with the trajectory. If the visual appearance of the non-drivable area is similar to the drivable area, the labeling process can't separate them. The most common case was detecting the sidewalk as drivable if the appearance was similar to the road (Figure \ref{struggle}c), resulting in lower performance in suburban driving. In some cases, the TADAP labels also identified the road shoulder as drivable in countryside driving if the drivable area boundary was not clear (Figure \ref{struggle}d). In these cases, the drivable area detections could be improved by measuring or predicting the 3D structure of the scene to detect the edges between drivable and non-drivable areas. For better scalability, structure from motion or neural radiance field approaches could be implemented instead of 3D lidar measurements.

GNSS-based trajectory extraction and projection to the image frame made with planarity assumption is easier to implement than 3D point cloud mapping-based methods at the expense of positioning and projection accuracy. Here, the projection error was found to be more significant than the positioning error but the GNSS positioning error could increase in densely built urban areas. The projection error increases when moving further ahead as the ground likely deviates more from the planarity assumption. Caused by the limited projection accuracy, the trajectory only included the following 50 m of driving instead of the 100 m used in the 3D point cloud mapping-based approaches. This can result in poor detection of drivable areas far ahead of the vehicle because the TADAP labels have difficulties detecting drivable areas that are far away from the trajectory (Figure \ref{struggle}a). For the same reason, the detection of adjacent lanes can also be challenging (Figure \ref{struggle}b). The trajectory estimate could be extended by using GNSS-INS fusion to estimate the height coordinate for a more accurate projection from the GNSS to the image frame. Utilizing 3D point cloud mapping would produce the most accurate trajectory estimate, but it makes the method less scalable due to the high price of 3D lidar sensors.   

The TADAP labels improve the existing trajectory-based methods by utilizing pre-trained self-supervised features for finding similar areas with the trajectory. Self-supervised features have been used in previous work, but instead of using pre-trained features, they try to learn suitable feature representations only from their data. Most pre-trained self-supervised vision models are trained with significantly larger datasets and model sizes, providing better generalization. On the other hand, our method solves the label assignment problem of the unsupervised segmentation models by using the mean feature of the trajectory as a reference point for the drivable area label.  

The datasets used in this paper included only winter driving, but the method should work in any driving conditions. Winter is one of the most challenging cases for drivable area detection, meaning it is likely that the TADAP labels could reach even higher performance in favorable weather conditions. This hypothesis is supported by the fact that the highway scene included the least snow in the drivable area and it had the highest performance. 

The presented TADAP auto-labeling process is extremely scalable as only a monocular camera and a GNSS unit are required. A large number of vehicles can be used for data collection with little cost, providing data from a variety of driving scenarios, locations, and conditions. Automated labeling enables the creation of significantly larger datasets than what is feasible with manual labeling. Based on this data, robust drivable area detection models that work in all driving conditions can be developed. 

\section*{Acknowledgments}
We express our gratitude to the Henry Ford Foundation Finland for providing funding for this research. We acknowledge the computational resources provided by the Aalto Science-IT project. We also want to thank Junwon Seo for his support in the implementation of their method. 

\bibliographystyle{IEEEtran}
\bibliography{ref}

\begin{IEEEbiography}[{\includegraphics[width=1in,height=1.25in,clip,keepaspectratio]{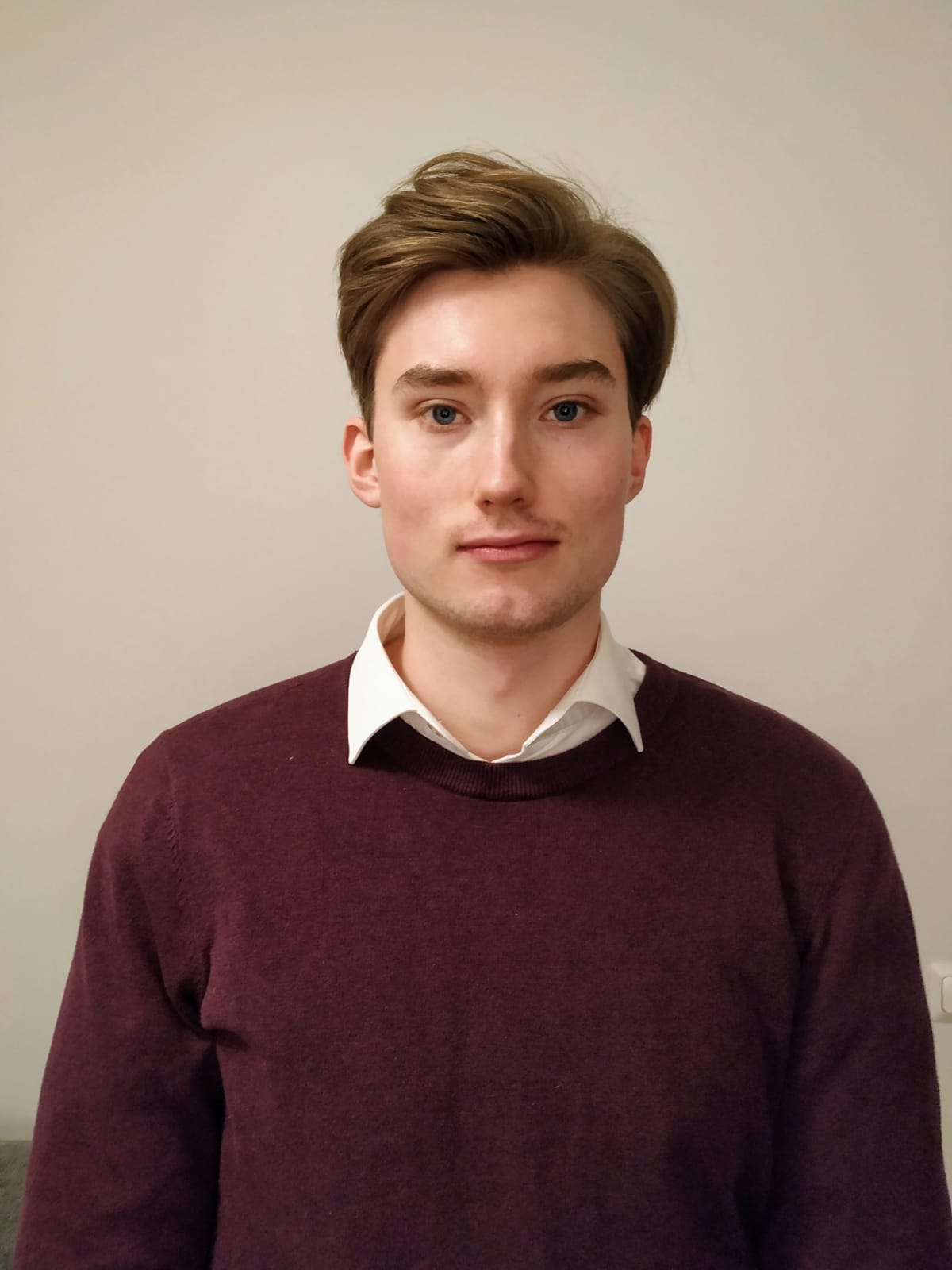}}]{Eerik Alamikkotervo}
received his BSc and MSc degrees from Aalto University in 2021 and 2023, respectively.
He is currently studying towards a DSc degree at Aalto University in the Autonomy \& Mobility laboratory.
His research interests are perception, autonomous mobile robotics, and machine learning. 
\end{IEEEbiography}

\begin{IEEEbiography}[{\includegraphics[width=1in,height=1.25in,clip,keepaspectratio]{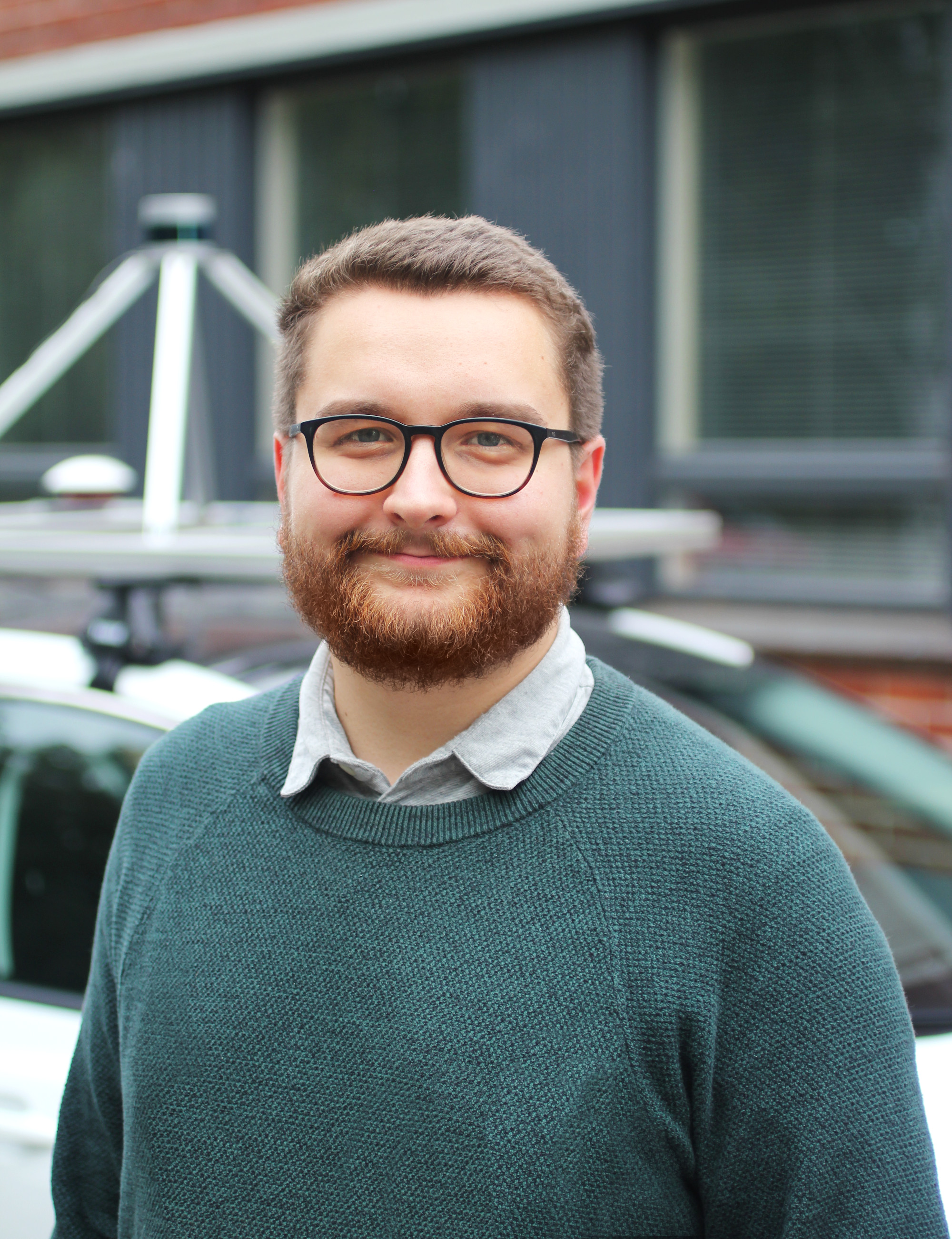}}]{Risto Ojala}
received his BSc, MSc, and DSc degrees from Aalto University in 2019, 2021, and 2023, respectively.
Currently, he continues at Aalto as a postdoctoral researcher at the Autonomy \& Mobility laboratory.
He has authored several peer-reviewed journal publications, and his research interests focus on automated vehicles, mobile robotics, computer vision, and machine learning.
\end{IEEEbiography}

\begin{IEEEbiography}[{\includegraphics[width=1in,height=1.25in,clip,keepaspectratio]{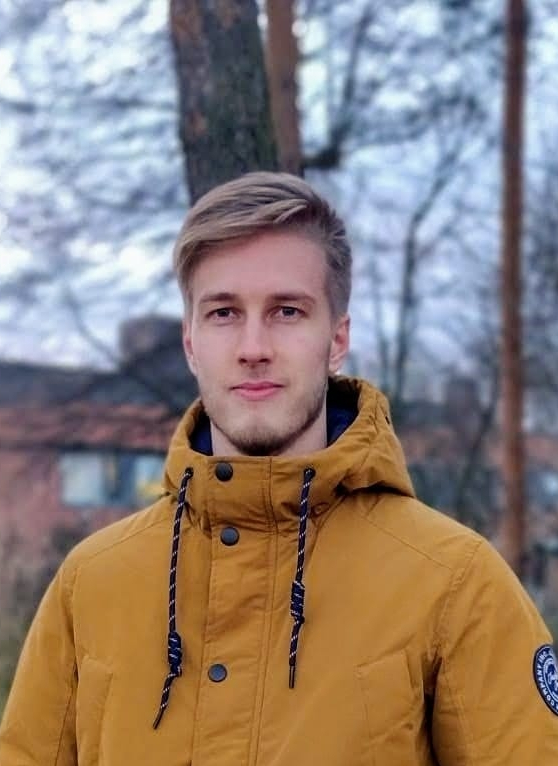}}]{Alvari Seppänen}
received his BSc and MSc degrees from Aalto University in 2020 and 2021, respectively.
He is currently studying towards a DSc degree at Aalto University in the Autonomy \& Mobility laboratory.
His research interests are mobile robotics and robot perception.
He has multiple peer-reviewed publications on these topics.
\end{IEEEbiography}

\begin{IEEEbiography}[{\includegraphics[width=1in,height=1.25in,clip,keepaspectratio]{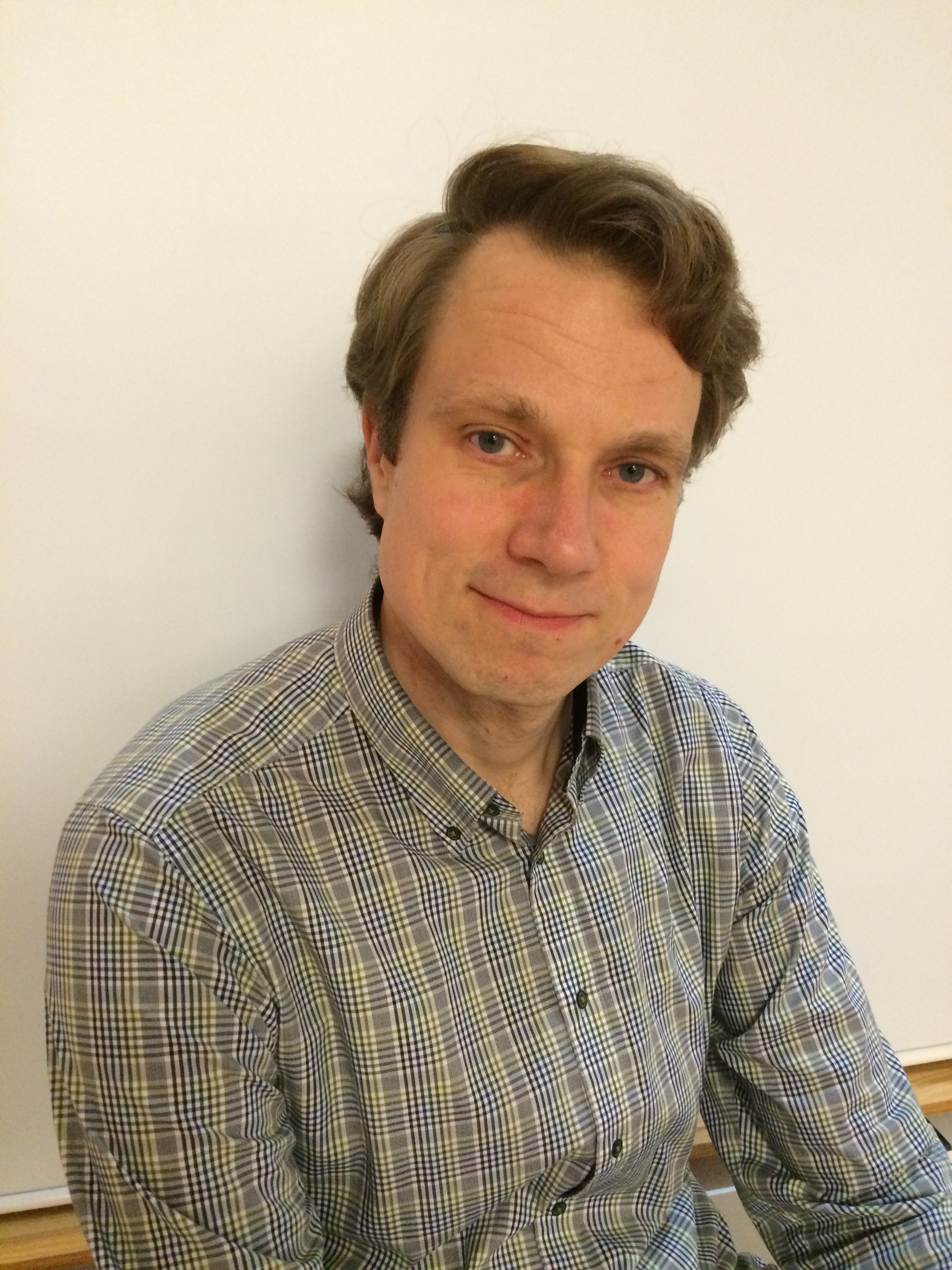}}]{Kari Tammi} 
has been a Professor at Aalto University since 2015. He currently works as Dean for the School of Engineering. He also serves the Finnish Administrative Supreme Court as a Chief Engineer Counselor. He received the M.Sc., Lic.Sc., and D.Sc. degrees from the Helsinki University of Technology, in 1999, 2003, and 2007, respectively. He received a Teacher’s pedagogical qualification at the Hame University of Applied Sciences, in 2017. He was a Researcher with CERN, the European Organization for Nuclear Research, from 1997 to 2000, and a Postdoctoral Researcher with North Carolina State University, USA, from 2007 to 2008. From 2000 to 2015, he held a Research Professor, a Research Manager, a Team Leader, and other positions at the VTT Technical Research Centre of Finland. He has authored over 120 peer-reviewed publications cited in over 7000 other publications. He is a member of the Finnish Academy of Technology and the President of The Association of Automotive Technical Societies in Finland.
\end{IEEEbiography}

\vfill

\end{document}